\documentclass[10pt,twocolumn,letterpaper]{article}

\usepackage{cvpr}              

\usepackage{graphicx}
\usepackage{amsmath}
\usepackage{amssymb}
\usepackage{booktabs}
\usepackage{csquotes} 
\newcommand{\xmark}{\normalsize \ding{55}}%
\usepackage{tabularx}
\usepackage{multirow}
\usepackage{amssymb}
\usepackage{pifont}
\usepackage{stfloats}
\usepackage{comment}

\usepackage[pagebackref,breaklinks,colorlinks]{hyperref}

\usepackage[capitalize]{cleveref}
\crefname{section}{Sec.}{Secs.}
\Crefname{section}{Section}{Sections}
\Crefname{table}{Table}{Tables}
\crefname{table}{Tab.}{Tabs.}

\def\cvprPaperID{*****} 
\def\confName{CVPR}
\def\confYear{2022}

\begin{document}

\title{MixAugment \& Mixup: \\ Augmentation Methods for Facial Expression Recognition}

\author{Andreas Psaroudakis\\
National Technical University of Athens, Greece\\
{\tt\small andreaspsaroudakis@gmail.com}
\and
Dimitrios Kollias\\
Queen Mary University of London, UK\\
{\tt\small d.kollias@qmul.ac.uk}
}
\maketitle

\begin{abstract}

Automatic Facial Expression Recognition (FER) has attracted increasing attention in the last 20 years since facial expressions play a central role in human communication. Most FER methodologies utilize Deep Neural Networks (DNNs) that are powerful tools when it comes to data analysis. However, despite their power, these networks are prone to overfitting, as they often tend to memorize the training data. What is more, there are not currently a lot of in-the-wild (i.e. in unconstrained environment) large databases for FER. To alleviate this issue, a number of data augmentation techniques have been proposed. Data augmentation is a way to increase the diversity of available data by applying constrained transformations on the original data. One such technique, which has positively contributed to various classification tasks, is Mixup. According to this, a DNN is trained on convex combinations of pairs of examples and their corresponding labels. 

In this paper, we examine the effectiveness of Mixup for in-the-wild FER in which data have large variations in head poses, illumination conditions, backgrounds and contexts. We then propose a new data augmentation strategy which is based on Mixup, called MixAugment. According to this, the network is trained concurrently on a combination of virtual examples and real examples; all these examples contribute to the overall loss function. We conduct an extensive experimental study that proves the effectiveness of MixAugment over Mixup and various state-of-the-art methods. We further investigate the combination of dropout with Mixup and MixAugment, as well as the combination of other data augmentation techniques with MixAugment.

\end{abstract}

\section{Introduction}
\label{sec:intro}

The human emotion constitutes a conscious subjective experience that can be expressed in various ways. During the past decade, with the rapid development in the field of Artificial Intelligence, scientists have conducted numerous studies to develop systems and robots that will be capable of perceiving automatically people’s feelings and behaviors \cite{kollias2016line,kollias2017adaptation}. An ultimate goal is the creation of digital assistants that will display a human-centered character and interact with users in the most natural way possible. It is a very complex and demanding task, since expression recognition in real world conditions is not easy and straightforward to do \cite{kollias2018training,kollias2015interweaving}.

Over the last decade, Deep Neural Networks have emerged as a method to solve any computer vision task \cite{tagaris1,tagaris2,mdpi}. DNNs in order to work and generalise well, need to be trained on large and diverse databases \cite{kollias2020exploiting,kollias2018old}. Nevertheless, in multiple applications, the collection of new data and their corresponding annotation is not always an easy or possible task to do (eg it is a quite time consuming and costly process). In the FER domain, RAFD-DB \cite{li2017reliable,li2019reliable}, AffectNet \cite{mollahosseini2017affectnet} and Aff-Wild2 \cite{kollias2022abaw,kollias2017recognition,kollias2019expression,kollias2019face,kollias2020analysing,kollias2021affect,kollias2021analysing,kollias2021distribution,zafeiriou2017aff,kollias2019deep,kollias2018multi,kollias2018aff} are the most widely used in-the-wild databases.
Additionally, despite DNNs' considerable power, the networks are prone to overfitting. This means that they often tend to memorize the input data or learn the noise and not the real data distribution, thus failing to generalize successfully when faced with data that are (considerably) different to the input ones.

One possible solution would be to expand the training set by adding new samples (although as we previously mentioned that is not always feasible due to inavailability of existing large in-the-wild datasets). Another way of extending the training set is by adding artificial samples that have been produced using 3D  methods \cite{fried2016perspective, averbuch2017bringing,thies2016face2face, kollias2018photorealistic,kollias2020deep} or Generative Adversarial Networks (GANs) \cite{goodfellow2014generative,zhou2017photorealistic,ding2018exprgan,pumarola2018ganimation,kollias2020va}. However, in this case, the generated samples must be realistic to the human eye, which still remains a very challenging task under investigation. In various application fields other problems may arise as well (eg for creating human faces the identity of the human should be preserved).

Another approach is to use data augmentation techniques, i.e., methods that produce new samples, by utilizing those that are already available and exist in the training set. Data augmentation is a way to increase the diversity of available data by applying constrained transformations on the original data.
A fairly recent technique of this kind, which has positively contributed to various tasks, is Mixup \cite{zhang2017mixup}. According to that, a DNN is trained on convex combinations of pairs of examples and their corresponding labels. By doing so, the distribution of the available data is extended and the generalization ability of the network improves. This principle has already been applied in some particular fields but has hardly ever been tried out in human affect estimation problems, especially in \enquote{in-the-wild} conditions with variations in head poses, illumination conditions, backgrounds and contexts.

In this paper, we examine the effectiveness of Mixup for 7 basic expression classification (categorical model \cite{ekman2002facial}) by utilizing the Real-world Affective Faces Database (RAF-DB) \cite{li2017reliable}, a large-scale facial expression database with around 30K great-diverse facial images downloaded from the Internet. We further  propose a new data augmentation strategy that is
based on Mixup, which we call MixAugment; according to this a DNN is trained  on a combination of virtual examples and real examples. The overall loss function of DNN training consists of the loss of the real examples and the loss of the virtual examples. Finally we examine the effect of dropout \cite{srivastava2014dropout} when used in combination with Mixup and our proposed MixAugment. Useful conclusions are drawn from the experimental study and the foundations are laid for future extensions.

\section{Related Work}

Mixup \cite{zhang2017mixup} constitutes a simple but powerful data augmentation routine that has already been applied in various tasks in Computer Vision, Natural Language Processing (NLP) and the audio domain. Some indicative examples pertain to medical image segmentation \cite{eaton2018improving}, sentence classification \cite{guo2019augmenting,chen2020mixtext,sun2020mixup}, audio tagging \cite{wei2018sample}, audio scene classification \cite{xu2018mixup} and image classification \cite{kollias2021mia,Tailor,kollias2020transparent,kollias2018deep1,hou2021cmc}.

Regarding expression recognition, Mixup has been tried out only in very limited scenarios. In particular, this data augmentation technique was applied for the first time in speech expression recognition (SER) data to alleviate the issue of small existing datasets in the field. In \cite{latif2020augmenting} a framework that combined Mixup with a Generative Adversarial Network was proposed so as to improve the generation of synthetic samples. Specifically, they utilized this routine to train a GAN for synthetic expression feature generation and also for learning expression feature representation. To prove the effectiveness of the proposed framework, they showed results for SER on synthetic feature vectors, augmentation of the training data with synthetic features and encoded features in compressed representation. The results indicated that the proposed network can successfully learn compressed expression representations and can also produce synthetic samples that enhance performance in within-corpus and cross-corpus evaluation. 

Apart from the lack of many large in-the-wild datasets in the field of SER, another problem is affiliated with the common difference between the training and test data distributions. SER systems can achieve high accuracy when these two sets are identically distributed, but this assumption is often violated in practice and the systems' performance declines against unforeseen data shifts. In \cite{latif2020deep}, the authors proved that the use of Mixup enhances the robustness to noise and adversarial examples in DNN Architectures. As a result, the generalization ability of the models improves and the DNNs perform better against unseen real-time situations. Moreover, the evaluations on the widely used IEMOCAP and MSP-IMPROV datasets showed that Mixup is a better augmentation technique for SER compared to the popular speed perturbation \cite{latif2019direct}.

Jia and Zheng \cite{jia2021emotion} tried to solve the problems of naturalness, robustness, fidelity and expression recognition accuracy in the process of expression speech synthesis. For that purpose, they designed an expression speech synthesis method based on multi-channel time–frequency generative adversarial networks (MC-TFD GANs) and Mixup. The comparative experiments were carried out on the IEMOCAP corpus. The results showed that the mean opinion score (MOS) and the unweighted accuracy (UA) of the speech generated by the synthesis method were improved by 4\% and 2.7\%, respectively. The proposed method was superior in subjective evaluation and objective experiments, proving that the speech produced by this model had higher reliability, better fluency and emotional expression ability.

In terms of expression recognition from facial images,a published work in which Mixup is utilized, is from  \cite{riaz2020exnet}. In this paper, the researchers made use of Mixup to improve the generalization of a proposed DNN, named eXnet. The model was trained and evaluated on FER-2013, CK+, and RAF-DB benchmark datasets. The experimental results showed that the model trained with Mixup technique witnessed an increase in accuracy of about 1\%. 


\section{Methodology}
\subsection{Mixup}
Mixup \cite{zhang2017mixup} is a simple and data-agnostic data augmentation routine that trains a DNN on convex combinations of pairs of examples and their labels. In other words, Mixup constructs virtual training examples $(\tilde{x},\tilde{y})$ as follows:
\begin{align}
\tilde{x} &= \lambda x_i+(1-\lambda)x_j  \nonumber \\ 
\tilde{y} &= \lambda y_i+(1-\lambda)y_j
\label{eq:mixup}
\end{align}

\noindent where $x_i$ and $x_j$ are two random raw inputs (i.e., images), $y_i$ and $y_j$ $\in \{0,1\}^7$ are their corresponding one-hot label encodings  and  $\lambda \thicksim \mathrm{B}(\alpha,\alpha) \in [0,1]$ (i.e., Beta distribution) for $\alpha \in (0, \infty)$.

Therefore, Mixup extends the training distribution by incorporating the prior knowledge that linear interpolations of feature vectors should lead to linear interpolations of the associated targets. By doing so, it regularizes the DNN (while training) to favor linear behavior in-between training examples. The implementation of Mixup training is straightforward, and introduces a minimal computation overhead. 

In the studied case, the training samples are aligned facial images and the labels are one-hot encoding vectors corresponding to one of the 7 basic expressions. When training a DNN with the Mixup technique, the mixup loss function is the categorical cross entropy (CCE) of the virtual (v) samples defined as:

\begin{align}
\mathcal{L}_{CCE}^{v} &=  \mathop{ \mathbb{E}}_{ \bar{y},\tilde{y}}[- \tilde{y} \cdot \text{log } \bar{y} ]   
\label{eq:cce}
\end{align}
where  $\bar{y}$  is the predicted probability of the sample $\tilde{x}$; $\tilde{x}$ and $\tilde{y}$ are given in Eq. \ref{eq:mixup}.

\begin{figure}[h!]
  \centering
 \includegraphics[scale=0.375]{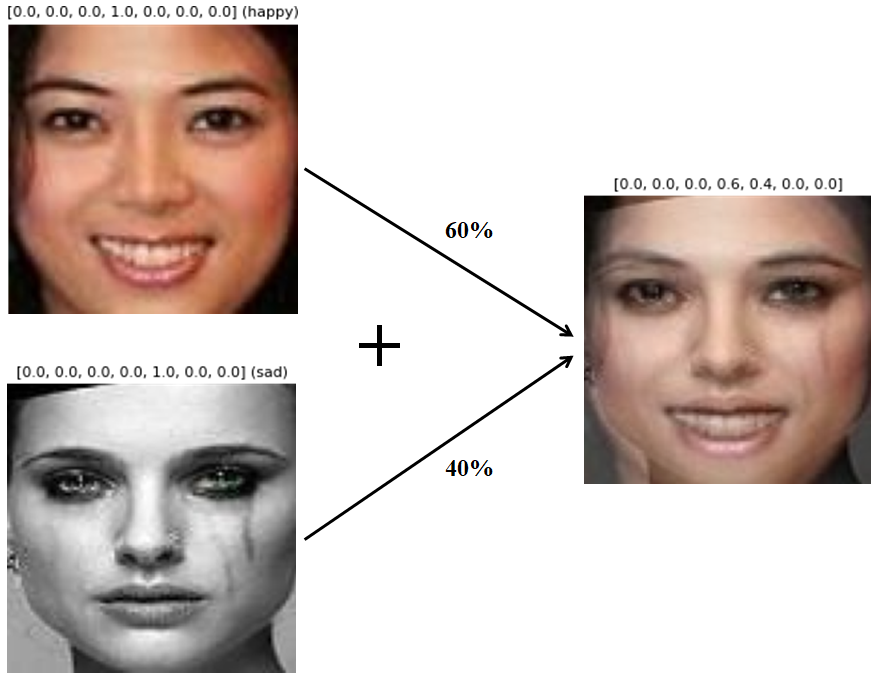}
 \caption{Construction of virtual example with the Mixup technique on samples taken from RAF-DB}
 \label{fig:mixup}
\end{figure}

An example of Mixup implementation on facial images is illustrated in Figure \ref{fig:mixup}. An image that corresponds to a \enquote{happy} facial expression is linearly mixed with another one that demonstrates a \enquote{sad} expression, in a 60:40 ratio. The resulting image depicts a human face, that combines facial characteristics from the two initial images. Its label, which is written above the constructed image, states that this virtual sample belongs to class \enquote{happy} by 60\% and to class \enquote{sad} by 40\%. 

\subsection{MixAugment}

\begin{figure}[h!]
  \centering
 \includegraphics[scale=0.375]{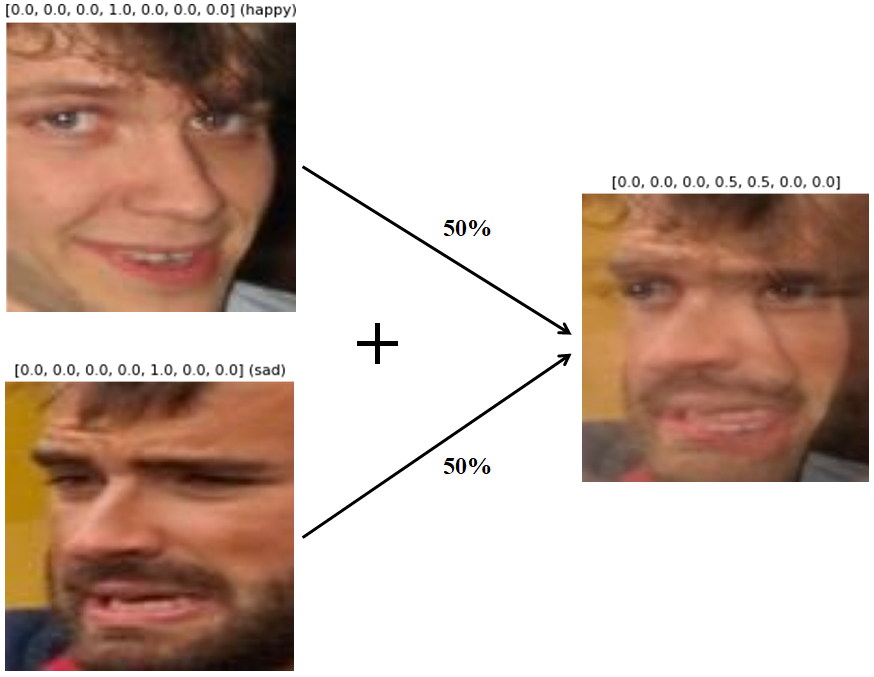}
 \caption{Example of mixing images with different head poses}
 \label{fig:MixAugment}
\end{figure}

In the typical Mixup data augmentation routine, randomly selected pairs of images are linearly interpolated and then fed into the DNN for training. However, \enquote{in-the-wild} facial databases contain a lot of images with large variations in head poses, gazes and angles. As a result, when mixing randomly selected images, it is possible for two images with different head poses to be combined. An indicative example is illustrated in Figure \ref{fig:MixAugment}, where a \enquote{happy} facial expression is mixed with a \enquote{sad} reaction, in a 50:50 ratio.

 As one can see, the resulting image does not resemble a real human face. Such cases may hinder training and learning of DNNs.
To cope with this problem, we propose a simple approach name MixAugment. According to this, during each training iteration, the DNN is trained concurrently on both real (r) and virtual (v) examples. Specifically, in each training iteration, the DNN is fed with both $x_i$ and $x_j$, and the generated image $\tilde{x} = \lambda x_i+(1-\lambda)x_j$ (of Eq. \ref{eq:mixup}). In this scenario, the loss function is: 
\begin{align}
&\mathcal{L}_{total} =  \mathcal{L}_{CCE}^{v} + \mathcal{L}_{CCE}^{r_i} + \mathcal{L}_{CCE}^{r_j} \nonumber   \\
&= \mathop{ \mathbb{E}}[- \tilde{y} \cdot \text{log } \bar{y} - y_i \cdot \text{log } \bar{y_i} - y_j \cdot \text{log } \bar{y_j} ] \nonumber   \\
& = \mathop{ \mathbb{E}} \big [- [\lambda y_i+(1-\lambda)y_j] \cdot \text{log } \bar{y} - y_i \cdot \text{log } \bar{y_i} - y_j \cdot \text{log } \bar{y_j} \big ] \nonumber   \\
&= \mathop{ \mathbb{E}}[- y_i \cdot \text{log } (\bar{y_i} \bar{y}^\lambda) - y_j \cdot \text{log } (\bar{y_j} \bar{y}^{1-\lambda}) ]   
\label{eq:proposed}
\end{align}
where  $\bar{y}$  is the predicted probability of the sample $\tilde{x}$; $\tilde{x}$ and $\tilde{y}$ are given in Eq. \ref{eq:mixup}; $y_i$ and $y_j$ are the labels of two (random) images (mentioned in Eq. \ref{eq:mixup}) and $\bar{y_i}$ and $\bar{y_j}$ are their corresponding predicted probabilities (the indices in the expectations are omitted for simplicity).

As can be seen in Eq. \ref{eq:proposed} we merge the mixup loss with the classification loss to enhance the classification ability on both raw samples and mixup samples. This is different from the original design of Mixup \cite{zhang2017mixup} where the authors replaced the classification loss with the mixup loss.

\section{Experimental Studies}

\subsection{Database}

\begin{figure*}[t]
  \centering
\resizebox{0.95\textwidth}{5cm}{
 \includegraphics{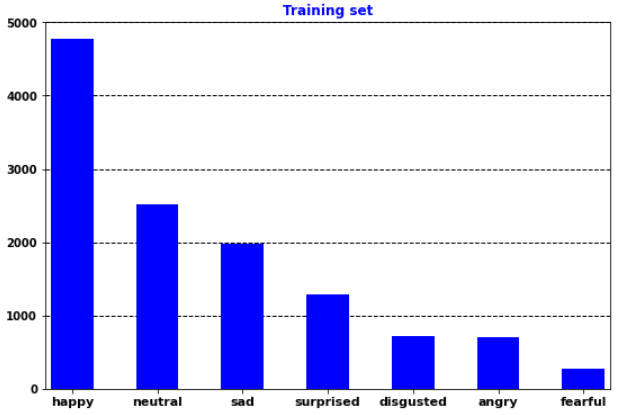} \hspace{1cm}
 \includegraphics{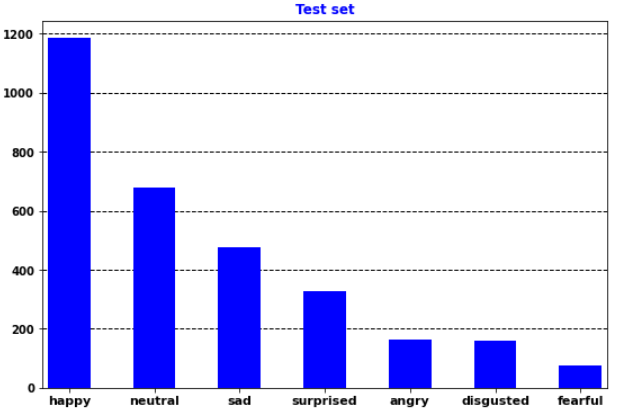}}
 \caption{Class distribution of the RAF-DB training and test sets}
 \label{fig:distribution}
\end{figure*}

All the experiments are carried out utilizing the Real-world Affective Faces Database (RAF-DB), a large-scale facial expression database with around 30K great-diverse facial images downloaded from the Internet. Based on the crowdsourcing annotation, each image has been independently labeled by about 40 annotators. Images in this database are of great variability in subjects' age, gender and ethnicity, head poses, lighting conditions, occlusions, (e.g. glasses, facial hair, self-occlusion). RAF-DB includes two subsets: (i) single-label subset, which consists of images annotated in terms of the seven basic expressions (surprise, fear, disgust, happiness, sadness, anger and neutral); (ii) multi-label subset, which consists of images annotated in terms of twelve compound expressions. In our experiments, we use the single-label subset. 
The database has been split into a training set (consisting of around 12,200 images) and a test set (consisting of around 3,100 images) where the size of training set is four times larger than the size of the test set; expressions in both sets have a near-identical distribution, as illustrated in Figure \ref{fig:distribution}. It is worth mentioning that the two sets are imbalanced, with the expression \enquote{happy} having by far the largest number of samples and the class \enquote{fearful} being the least popular in both cases.

\subsection{Performance Metric}
For the evaluation of our models we make use of three different performance metrics: i) Accuracy, ii) Average Accuracy (mean diagonal of the normalized confusion matrix) and iii) macro F1-score (harmonic mean of Precision and Recall). Accuracy is defined as the percentage of correct predictions among the total number of predictions. It is the most common evaluation metric, however not preferred in imbalanced classification problems. Our dataset is imbalanced, therefore, to have a superior insight, we should take into account some additional metrics. The Average Accuracy (aka macro Recall) and macro F1-score are useful, since they give equal importance to each class, in contrast to Accuracy which  gives equal importance to each sample, thus favoring majority classes. During the training phase, we monitor these metrics, and if no improvement is observed over the test set, we apply early stopping and keep the best configuration.

\subsection{Pre-processing}

Data pre-processing consists of the steps required for facilitating extraction of meaningful features from the data. In a typical expression recognition problem with facial images, the usual steps are face detection and alignment, image resizing and image normalization. We experimented with using two different face detectors to extract bounding boxes around each face and detect 68 facial landmarks. In the first version of the database (the public release), a face detector from the dlib library has been used, while in the other case the detector is the RetinaFace \cite{deng2020retinaface}. The alignment step is the same for both versions. Out of all 68 located landmarks, we focus on 5 - corresponding to the location of the left eye, right eye, nose and mouth in a prototypical frontal face - as rigid, anchor points. Then, for every image, the respective 5 facial landmarks are extracted and affinity transformations between the coordinates of these 5 landmarks and the coordinates of the 5 landmarks of the frontal face are computed; these transformations are imposed to the whole new frame for the alignment to be performed. All resulting images are resized to $100 \times 100 \times 3$ or $112 \times 112 \times 3$. Finally, all cropped and aligned images' pixel intensity values are normalized to the range $[0,1]$.

\subsection{Training Implementation Details}

Table \ref{tab:parameters} demonstrates all implementation details pertained to the training session. Where dropout \cite{srivastava2014dropout} was applied, its value was 0.5. In the following, to not clutter the presented results, we present results only for the publicly-released dataset version (1st version); the same conclusions have been drawn when utilizing the other version (2nd).

\begin{table}[h]
\caption{Training parameters with their corresponding values}
\setlength{\tabcolsep}{12pt}
\centering
\resizebox{\linewidth}{!}{
\begin{tabular}{cc}
\hline
\textbf{Parameters} & \textbf{Values}                                                                                   \\ \hline \hline 
Image size          & \begin{tabular}[c]{@{}c@{}}$100 \times 100 \times 3$ (1st version)\\ $112 \times 112 \times 3$ (2nd version)\end{tabular} \\ 
Batch size          & $32$                                                                                                \\ 
Loss function           & Categorical cross entropy                                  \\ 
Optimizer           & Adam                                                                                              \\ 
Learning rate       & $10^{-3}, 10^{-4}$                                                                            \\ 
Dropout rate      & $0.5$                                                                            \\ 
Number of epochs    & $100$      
\\ \hline
\end{tabular}}
\label{tab:parameters}
\end{table}

\subsection{Results}


\begin{table}[h!]
\caption{ResNet50 trained with Mixup and without Mixup (i.e., vanilla case)}
\setlength{\tabcolsep}{3.5pt}
\centering
\resizebox{\linewidth}{!}{
\begin{tabular}{ccccc}
\hline
\multicolumn{1}{c}{\textbf{Mixup}}           & \multicolumn{1}{c}{\textbf{Dropout}} & \multicolumn{1}{c}{\textbf{Accuracy}} & \multicolumn{1}{c}{\textbf{F1-score}} & \textbf{Aver. Acc.} \\ \hline \hline
\multicolumn{1}{c}{\multirow{2}{*}{No}} & \multicolumn{1}{c}{\xmark}               & \multicolumn{1}{c}{83,31}                  & \multicolumn{1}{c}{75,25}    & 73,46          \\ 
\multicolumn{1}{c}{}                         & \multicolumn{1}{c}{\large \checkmark}              & \multicolumn{1}{c}{83,21}                  & \multicolumn{1}{c}{75,16}        & 74,07                      \\ \hline 
\multicolumn{1}{c}{\multirow{2}{*}{\boldmath{$\alpha = 0.1$}}} & \multicolumn{1}{c}{\xmark}               & \multicolumn{1}{c}{\textbf{84,06}}                  & \multicolumn{1}{c}{\textbf{75,51}}                         & \textbf{74,38}     \\  
\multicolumn{1}{c}{}                         & \multicolumn{1}{c}{\large \checkmark}              & \multicolumn{1}{c}{83,25}                  & \multicolumn{1}{c}{74,88}                   & 73,60   \\ \hline
\multicolumn{1}{c}{\multirow{2}{*}{$\alpha = 0.2$}} & \multicolumn{1}{c}{\xmark}               & \multicolumn{1}{c}{82,33}                  & \multicolumn{1}{c}{73,95}                 & 73,38       \\  
\multicolumn{1}{c}{}                         & \multicolumn{1}{c}{\large \checkmark}              & \multicolumn{1}{c}{83,12}                  & \multicolumn{1}{c}{74,62}          & 74,14                    \\ \hline \multicolumn{1}{c}{\multirow{2}{*}{$\alpha = 0.6$}} & \multicolumn{1}{c}{\xmark}               & \multicolumn{1}{c}{83,15}                  & \multicolumn{1}{c}{74,81}                          & 72,73    \\  
\multicolumn{1}{c}{}                         & \multicolumn{1}{c}{\large \checkmark}              & \multicolumn{1}{c}{83,47}                  & \multicolumn{1}{c}{75,23}                       & 74,29        \\ \hline
\multicolumn{1}{c}{\multirow{2}{*}{$\alpha = 1$}} & \multicolumn{1}{c}{\xmark}               & \multicolumn{1}{c}{82,55}                  & \multicolumn{1}{c}{74,01}                            & 73,29   \\  
\multicolumn{1}{c}{}                         & \multicolumn{1}{c}{\large \checkmark}              & \multicolumn{1}{c}{82,67}                  & \multicolumn{1}{c}{74,69}                          & 73,58     \\ \hline
\end{tabular}}
\label{tab:mixup}
\end{table}

\paragraph{Utilize Mixup vs Vanilla Case} We start our experiments by training a ResNet50 \cite{he2016deep}, pretrained on ImageNet, for 100 epochs, when applying and not applying dropout (vanilla case). We also train the exact same model (same weight initialization) with Mixup, for $\alpha \in \{0.1, 0.2, 0.6, 1\}$. Table \ref{tab:mixup} illustrates the results of these experiments. Let us not that  large values of the hyperparameter $(\alpha \in \{4,8\})$ in Mixup lead to underfitting and not good network performance. In Table \ref{tab:mixup}, it can be seen that the best configuration (i.e., when the model trained with Mixup achieves its higher performance in all 3 studied metrics) is  when $\alpha=0.1$ and no dropout has been used. In this case the model trained with Mixup outperforms by at least 0.3\% in all studied evaluation metrics the model trained without Mixup. In terms of the use of dropout, Table \ref{tab:mixup} shows that its addition sometimes contributes positively, whereas some other times seems to contribute negatively.

\paragraph{Utilize MixAugment vs Vanilla Case} Similar as before, we use the same model (ResNet50 pretrained on ImageNet) with the same training parameters and compare its performance when the proposed MixAugment is and is not used. Table \ref{tab:MixAugment} illustrates that performance comparison for $\alpha \in \{0.1, 0.2, 0.6, 1\}$ and when dropout is and is not applied. Similarly as in the Mixup case, we noticed that for large values of the hyperparameter $(\alpha \in \{4,8\})$ Mixup leads to underfitting and and not good network performance. In Table \ref{tab:MixAugment}, it can be seen that the best configuration is when $\alpha=0.1$ and no dropout has been used. In this case, the model trained with MixAugment outperforms by at least 1.7\% in all studied evaluation metrics the model trained without MixAugment. In terms of the use of dropout, Table \ref{tab:MixAugment} shows that its addition sometimes contributes positively, whereas some other times seems to contribute negatively. 
Finally, one can observe in Tables  \ref{tab:mixup}  and  \ref{tab:MixAugment}, that in both cases (when Mixup or MixAugment have been used), best results across all metrics have been obtained when $\alpha=0.1$ and no dropout has been used. We can deduct that optimal results cannot be achieved when dropout is used in addition to Mixup or MixAugment.

\begin{table}[h!]
\caption{ResNet50 trained with the proposed MixAugment and without MixAugment (i.e., vanilla case)}
\setlength{\tabcolsep}{3pt}
\centering
\resizebox{\linewidth}{!}{
\begin{tabular}{ccccc}
\hline
\multicolumn{1}{c}{\textbf{MixAugment}}           & \multicolumn{1}{c}{\textbf{Dropout}} & \multicolumn{1}{c}{\textbf{Accuracy}} & \multicolumn{1}{c}{\textbf{F1-score}} & \textbf{Aver. Acc.} \\ \hline \hline
\multicolumn{1}{c}{\multirow{2}{*}{No}} & \multicolumn{1}{c}{\xmark}               & \multicolumn{1}{c}{83,31}                  & \multicolumn{1}{c}{75,25}    & 73,46          \\ 
\multicolumn{1}{c}{}                         & \multicolumn{1}{c}{\large \checkmark}              & \multicolumn{1}{c}{83,21}                  & \multicolumn{1}{c}{75,16}        & 74,07                      \\ \hline 
\multicolumn{1}{c}{\multirow{2}{*}{\boldmath{$\alpha = 0.1$}}} & \multicolumn{1}{c}{\xmark}               & \multicolumn{1}{c}{\textbf{85,04}}                  & \multicolumn{1}{c}{\textbf{77,30}}                         & \textbf{75,32}     \\  
\multicolumn{1}{c}{}                         & \multicolumn{1}{c}{\large \checkmark}              & \multicolumn{1}{c}{83,96}                  & \multicolumn{1}{c}{76,03}                   & 73,77   \\ \hline
\multicolumn{1}{c}{\multirow{2}{*}{$\alpha = 0.2$}} & \multicolumn{1}{c}{\xmark}               & \multicolumn{1}{c}{84,19}                  & \multicolumn{1}{c}{76,57}                 & 74,74       \\  
\multicolumn{1}{c}{}                         & \multicolumn{1}{c}{\large \checkmark}              & \multicolumn{1}{c}{84,39}                  & \multicolumn{1}{c}{76,64}          & 75,13                    \\ \hline 
\multicolumn{1}{c}{\multirow{2}{*}{$\alpha = 0.6$}} & \multicolumn{1}{c}{\xmark}               & \multicolumn{1}{c}{84,13}                  & \multicolumn{1}{c}{75,04}                      & 73,38   \\  
\multicolumn{1}{c}{}                         & \multicolumn{1}{c}{\large \checkmark}              & \multicolumn{1}{c}{84,26}                  & \multicolumn{1}{c}{76,46}                   & 74,38           \\ \hline 
\multicolumn{1}{c}{\multirow{2}{*}{$\alpha = 1$}} & \multicolumn{1}{c}{\xmark}               & \multicolumn{1}{c}{83,74}                  & \multicolumn{1}{c}{75,43}                          & 73,87     \\  
\multicolumn{1}{c}{}                         & \multicolumn{1}{c}{\large \checkmark}              & \multicolumn{1}{c}{83,51}                  & \multicolumn{1}{c}{75,58}                       & 74,36        \\ \hline
\end{tabular}}
\label{tab:MixAugment}
\end{table}

Next, in Table \ref{tab:confidences} we present the model's confidence for the correct and wrong predictions under two settings: i) when the model is trained with the proposed MixAugment and ii)  when it is trained without MixAugment (this is the vanilla case).
As illustrated in Table \ref{tab:confidences}, our proposed technique helps the network make right decisions with higher confidence and wrong decisions with less assuredness, which is obviously desirable. 


\begin{table}[h]
\caption{Prediction confidence with MixAugment and without MixAugment (i.e. vanilla case)}
\centering
\setlength{\tabcolsep}{3pt}
\resizebox{\linewidth}{!}{
\begin{tabular}{ccccc}
\hline
\textbf{Confidence}      & \multicolumn{2}{c}{\textbf{mean}}            & \multicolumn{2}{c}{\textbf{median}}          \\ \hline
Type of predictions & \multicolumn{1}{c}{correct} & wrong & \multicolumn{1}{c}{correct} & wrong \\ \hline \hline 
No MixAugment (Vanilla)             & \multicolumn{1}{c}{96,37}   & 92,66 & \multicolumn{1}{c}{98,82}   & 99,69 \\ 
\textbf{MixAugment}            & \multicolumn{1}{c}{\textbf{98,77}}   & \textbf{84,24} & \multicolumn{1}{c}{\textbf{100,0}}   & \textbf{90,75} \\ \hline
\end{tabular}}
\label{tab:confidences}
\end{table}

In addition, it would be interesting to examine the performance of MixAugment  separately on each one of the 7 basic expression categories across various metrics. For that purpose, in Table \ref{tab:classes}, we present the Precision, Recall and F1-score for each class when the model is trained with and without MixAugment. It can be seen that when the network is trained with MixAugment, in the vast majority of the categories, there is a substantial rise in all the aforementioned metrics. It is worth mentioning that the highest growth is observed for the minority class \enquote{fearful}. Particularly, Precision, Recall and F1-score increase by approximately 10\%, 5\% and 6\% respectively when MixAugment is used. This is a really promising result, since there are many classification tasks in which the classes with the smallest number of samples are of paramount importance (e.g. medical classification problems).

\begin{table*}[t]
\caption{Comparison between ResNet50 trained with the proposed MixAugment and without MixAugment (i.e., vanilla case) for each one of the 7 classes}
\centering
\setlength{\tabcolsep}{5pt}
\resizebox{0.97\textwidth}{!}{
\begin{tabular}{c|cc|cc|cc|c}
\hline
                     \multirow{2}{*}{\textbf{Class}}    & \multicolumn{2}{c}{\textbf{Precision}} & \multicolumn{2}{c}{\textbf{Recall}}  & \multicolumn{2}{c|}{\textbf{F1-score}} & \multirow{2}{*}{\textbf{Samples}} \\ \cline{2-7}
           & \begin{tabular}{@{}c@{}} No MixAugment \\ (Vanilla) \end{tabular}    & MixAugment  & \begin{tabular}{@{}c@{}} No MixAugment \\ (Vanilla) \end{tabular}   & MixAugment & \begin{tabular}{@{}c@{}} No MixAugment \\ (Vanilla) \end{tabular}    & MixAugment &                  \\ \hline \hline
surprised                 & \multicolumn{1}{c}{83,49}  & $\uparrow$ \textbf{85,14}     & \multicolumn{1}{c}{79.94} & $\uparrow$ \textbf{83,59}    & \multicolumn{1}{c}{81,68}  & $\uparrow$ \textbf{84,36}    & 329              \\ \hline
\textcolor{blue}{\textbf{fearful}}                   & \multicolumn{1}{c}{69,09}  & $\uparrow$ \textcolor{blue}{\textbf{78,85}}     & \multicolumn{1}{c}{51,35} & $\uparrow$ \textcolor{blue}{\textbf{55,41}}    & \multicolumn{1}{c}{58,91}  & $\uparrow$ \textcolor{blue}{\textbf{65,08}}    & \textcolor{blue}{\textbf{74}}              \\ \hline
disgusted                 & \multicolumn{1}{c}{62,60}  & $\uparrow$ \textbf{65,89}     & \multicolumn{1}{c}{51,25} & $\uparrow$ \textbf{53,12}    & \multicolumn{1}{c}{56,36}  & $\uparrow$ \textbf{58,82}    & 160              \\ \hline
happy                     & \multicolumn{1}{c}{92,33}  & $\uparrow$ \textbf{92,73}     & \multicolumn{1}{c}{93,50} & $\downarrow$ 92,57    & \multicolumn{1}{c}{92,91}  & $\downarrow$ 92,65    & 1185              \\ \hline
sad                       & \multicolumn{1}{c}{81,01}  & $\uparrow$ \textbf{83,37}     & \multicolumn{1}{c}{80,33} & $\uparrow$ \textbf{82,85}    & \multicolumn{1}{c}{80,67}  & $\uparrow$ \textbf{83,11}    & 478              \\ \hline
angry                     & \multicolumn{1}{c}{79,47}  & $\downarrow$ 75,08     & \multicolumn{1}{c}{74,07} & $\downarrow$ 71,60    & \multicolumn{1}{c}{76,68}  & $\downarrow$ 
75,08    & 162              \\ \hline
neutral                   & \multicolumn{1}{c}{78,30}  & $\downarrow$ 77,73     & \multicolumn{1}{c}{85,44} & $\uparrow$ \textbf{86,76}    & \multicolumn{1}{c}{81,72}  & $\uparrow$ \textbf{82,00}    & 680              \\ \hline 
\end{tabular}}
\label{tab:classes}
\end{table*}


\paragraph{Utilize MixAugment vs Utilize Mixup vs Vanilla Case}  To summarise the main presented results and to illustrate the difference in ResNet50's performance when the model is trained with the proposed MixAugment, when it is trained with Mixup and when it is trained without any of these, we have created Table \ref{sum}. It can be observed that Mixup improves the model's performance and MixAugment further improves its performance. Compared to Mixup, our technique further improves all three evaluation metrics for at least 1\%. Finally, it is notable to mention that when MixAugment is used in network training, the convergence is faster compared to the cases when Mixup is used or when neither of the two is used.

\begin{table}[h!]
\setlength{\tabcolsep}{8pt}
\caption{ResNet50 trained with MixAugment, with Mixup and without any of the two}
\label{sum}
\centering
\resizebox{\linewidth}{!}{
\begin{tabular}{ cccc }
\hline
 \textbf{Method} & \textbf{Accuracy} & \textbf{F1-score} & \textbf{Aver. Acc.} \\
\hline
\hline
Vanilla &  83,31 & 75,25 & 73,46 \\
\hline
Mixup   &  84,06 & 75,51 & 74,38  \\
\hline
MixAugment   & \textbf{85,04} & \textbf{77,30} & \textbf{75,32}  \\
 \hline
\begin{tabular}{@{}c@{}} MixAugment \\ + Flipping \end{tabular}  & \textbf{86,06} & \textbf{78,24} & \textbf{76,28}  \\
 \hline
\end{tabular}}
\end{table}

Finally, let us mention a final experiment that we conducted. When training the model (ResNet50) with the proposed MixAugment we further performed flipping, which resulted in further boosting the model's performance by around 1\% in each studied metric (Accuracy, F1-score and Average Accuracy).

\paragraph{Utilize MixAugment with other DNNs} We further used our proposed MixAugment when training other widely used DNNs, such as VGG16 \cite{simonyan2014very}, DenseNet121 \cite{huang2017densely} and EfficientNet \cite{tan2019efficientnet}. We noticed the same observations as before (i.e., as in the case of ResNet50 described previously). In more detail, the performance of these  networks trained with MixAugment outperformed -in all 3 studied metrics- the performance of the networks trained with Mixup, which outperformed -over all metrics- the performance of the corresponding vanilla networks.

\paragraph{Utilize MixAugment vs State-of-the-Art} In the previous cases, our model (ResNet50) was only pre-trained on ImageNet. It is known that if the model is further pre-trained on a similar task to the studied one (which is FER), then its performance further increases. Therefore we first pre-trained ResNet50 on AffectNet and then trained it with MixAugment and flipping. In Table \ref{sota} we compare its performance to the performance of various state-of-the-art methods. 
It can be observed that our approach outperforms all state-of-the-art methods in the accuracy metric and shows a slightly worse performance than two state-of-the-art methods (FaceBehaviorNet (Residual) \cite{kollias2021distribution}) and VGGFACE \cite{kollias2020deep}) in the average accuracy metric.

\begin{table}[h!]
\caption{Performance comparison between state-of-the-art methods and ResNet50 trained with MixAugment}
\label{sota}
\centering
\scalebox{0.92}{
\begin{tabular}{ ccc }
\hline
 \textbf{Method} & \textbf{Accuracy} & \textbf{Aver. Acc.} \\
\hline
\hline
RAN \cite{wang2020region} &  86,90 & - \\
\hline
Ad-Corre \cite{fard2022ad} &  86,96 & - \\
\hline
mSVM + DLP-CNN \cite{li2017reliable,li2019reliable}  &  - & 74,20 \\
\hline
MT-ArcRes \cite{kollias2019expression} &  - & 75,00 \\
 \hline
MT-ArcVGG \cite{kollias2019expression} &  - & 76,00 \\
 \hline
FaceBehaviorNet (VGG) \cite{kollias2019face} &  - & 71,00 \\
  \hline
FaceBehaviorNet (Residual) \cite{kollias2021distribution} &  - & \textbf{78,00} \\
  \hline
VGGFACE \cite{kollias2020deep} &  - & 77,50 \\
  \hline  
\begin{tabular}{@{}c@{}} pre-train, MixAugment + Flipping \end{tabular}  & \textbf{87,54}  & 77,30  \\
 \hline
\end{tabular}
}
\end{table} 

 The FaceBehaviorNet (Residual) \cite{kollias2021distribution} is a multi-task learning network that has been trained on over 5M of images. The VGGFACE \cite{kollias2020deep} is a network that has been trained on an augmented training set consisting of the training set of RAF-DB plus 13,000 other synthetic/generated images (more than the training size of RAF-DB); therefore our method is expected to perform worse than such methods.

\section{Conclusion and Future Work}

In this paper, at first, we examine the effectiveness of
Mixup for in-the-wild Facial Expression Recognition. Mixup is a data augmentation technique in which a DNN is trained on convex
combinations of pairs of examples and their corresponding labels. Taking into account that in in-the-wild FER people display high variations in head poses, illumination conditions, backgrounds and contexts, we have proposed a variation of Mixup, called MixAugment in which the network is trained on a combination of virtual examples (generated by Mixup) and real examples; in our approach both examples contribute to the overall loss function. 
We have conducted a large experimental study that includes: performance comparison between models trained with Mixup, MixAugment or without any of the two versus state-of-the-art methods; ablation studies; testing the combination of Mixup or MixAugment and dropout; testing the combination of MixAugment and other data augmentation techniques such as flipping. The experimental study proves that models perform better when our proposed MixAugment is used during training.


In our future work, we aim to extend and apply the proposed MixAugment technique to other small and large-scale \enquote{in-the-wild} datasets, as well as to other affect recognition tasks, such as valence-arousal estimation and action unit detection.

{\small
\bibliographystyle{ieee_fullname}
\bibliography{egbib}

\begin{thebibliography}{10}\itemsep=-1pt

\bibitem{averbuch2017bringing}
Hadar Averbuch-Elor, Daniel Cohen-Or, Johannes Kopf, and Michael~F Cohen.
\newblock Bringing portraits to life.
\newblock {\em ACM Transactions on Graphics (TOG)}, 36(6):196, 2017.

\bibitem{chen2020mixtext}
Jiaao Chen, Zichao Yang, and Diyi Yang.
\newblock Mixtext: Linguistically-informed interpolation of hidden space for
  semi-supervised text classification.
\newblock {\em arXiv preprint arXiv:2004.12239}, 2020.

\bibitem{deng2020retinaface}
Jiankang Deng, Jia Guo, Evangelos Ververas, Irene Kotsia, and Stefanos
  Zafeiriou.
\newblock Retinaface: Single-shot multi-level face localisation in the wild.
\newblock In {\em Proceedings of the IEEE/CVF Conference on Computer Vision and
  Pattern Recognition}, pages 5203--5212, 2020.

\bibitem{ding2018exprgan}
Hui Ding, Kumar Sricharan, and Rama Chellappa.
\newblock Exprgan: Facial expression editing with controllable expression
  intensity.
\newblock In {\em Thirty-Second AAAI Conference on Artificial Intelligence},
  2018.

\bibitem{eaton2018improving}
Zach Eaton-Rosen, Felix Bragman, Sebastien Ourselin, and M~Jorge Cardoso.
\newblock Improving data augmentation for medical image segmentation.
\newblock 2018.

\bibitem{ekman2002facial}
Paul Ekman.
\newblock Facial action coding system (facs).
\newblock {\em A human face}, 2002.

\bibitem{fard2022ad}
Ali~Pourramezan Fard and Mohammad~H Mahoor.
\newblock Ad-corre: Adaptive correlation-based loss for facial expression
  recognition in the wild.
\newblock {\em IEEE Access}, 10:26756--26768, 2022.

\bibitem{fried2016perspective}
Ohad Fried, Eli Shechtman, Dan~B Goldman, and Adam Finkelstein.
\newblock Perspective-aware manipulation of portrait photos.
\newblock {\em ACM Transactions on Graphics (TOG)}, 35(4):128, 2016.

\bibitem{goodfellow2014generative}
Ian Goodfellow, Jean Pouget-Abadie, Mehdi Mirza, Bing Xu, David Warde-Farley,
  Sherjil Ozair, Aaron Courville, and Yoshua Bengio.
\newblock Generative adversarial nets.
\newblock {\em Advances in neural information processing systems}, 27, 2014.

\bibitem{guo2019augmenting}
Hongyu Guo, Yongyi Mao, and Richong Zhang.
\newblock Augmenting data with mixup for sentence classification: An empirical
  study.
\newblock {\em arXiv preprint arXiv:1905.08941}, 2019.

\bibitem{he2016deep}
Kaiming He, Xiangyu Zhang, Shaoqing Ren, and Jian Sun.
\newblock Deep residual learning for image recognition.
\newblock In {\em Proceedings of the IEEE conference on computer vision and
  pattern recognition}, pages 770--778, 2016.

\bibitem{hou2021cmc}
Junlin Hou, Jilan Xu, Rui Feng, Yuejie Zhang, Fei Shan, and Weiya Shi.
\newblock Cmc-cov19d: Contrastive mixup classification for covid-19 diagnosis.
\newblock In {\em Proceedings of the IEEE/CVF International Conference on
  Computer Vision}, pages 454--461, 2021.

\bibitem{huang2017densely}
Gao Huang, Zhuang Liu, Laurens Van Der~Maaten, and Kilian~Q Weinberger.
\newblock Densely connected convolutional networks.
\newblock In {\em Proceedings of the IEEE conference on computer vision and
  pattern recognition}, pages 4700--4708, 2017.

\bibitem{jia2021emotion}
Ning Jia and Chunjun Zheng.
\newblock Emotion speech synthesis method based on multi-channel
  time--frequency domain generative adversarial networks (mc-tfd gans) and
  mixup.
\newblock {\em Arabian Journal for Science and Engineering}, pages 1--14, 2021.

\bibitem{kollias2022abaw}
Dimitrios Kollias.
\newblock Abaw: Valence-arousal estimation, expression recognition, action unit
  detection \& multi-task learning challenges.
\newblock {\em arXiv preprint arXiv:2202.10659}, 2022.

\bibitem{kollias2021mia}
Dimitrios Kollias, Anastasios Arsenos, Levon Soukissian, and Stefanos Kollias.
\newblock Mia-cov19d: Covid-19 detection through 3-d chest ct image analysis.
\newblock In {\em Proceedings of the IEEE/CVF International Conference on
  Computer Vision}, pages 537--544, 2021.

\bibitem{Tailor}
Dimitrios Kollias, N Bouas, Y Vlaxos, V Brillakis, M Seferis, Ilianna Kollia,
  Levon Sukissian, James Wingate, and S Kollias.
\newblock Deep transparent prediction through latent representation analysis.
\newblock {\em arXiv preprint arXiv:2009.07044}, 2020.

\bibitem{kollias2018photorealistic}
Dimitrios Kollias, Shiyang Cheng, Maja Pantic, and Stefanos Zafeiriou.
\newblock Photorealistic facial synthesis in the dimensional affect space.
\newblock In {\em Proceedings of the European Conference on Computer Vision
  (ECCV) Workshops}, pages 0--0, 2018.

\bibitem{kollias2020deep}
Dimitrios Kollias, Shiyang Cheng, Evangelos Ververas, Irene Kotsia, and
  Stefanos Zafeiriou.
\newblock Deep neural network augmentation: Generating faces for affect
  analysis.
\newblock {\em International Journal of Computer Vision}, 128(5):1455--1484,
  2020.

\bibitem{kollias2015interweaving}
Dimitris Kollias, George Marandianos, Amaryllis Raouzaiou, and Andreas-Georgios
  Stafylopatis.
\newblock Interweaving deep learning and semantic techniques for emotion
  analysis in human-machine interaction.
\newblock In {\em 2015 10th International Workshop on Semantic and Social Media
  Adaptation and Personalization (SMAP)}, pages 1--6. IEEE, 2015.

\bibitem{kollias2017recognition}
Dimitrios Kollias, Mihalis~A Nicolaou, Irene Kotsia, Guoying Zhao, and Stefanos
  Zafeiriou.
\newblock Recognition of affect in the wild using deep neural networks.
\newblock In {\em Computer Vision and Pattern Recognition Workshops (CVPRW),
  2017 IEEE Conference on}, pages 1972--1979. IEEE, 2017.

\bibitem{kollias2020analysing}
D Kollias, A Schulc, E Hajiyev, and S Zafeiriou.
\newblock Analysing affective behavior in the first abaw 2020 competition.
\newblock In {\em 2020 15th IEEE International Conference on Automatic Face and
  Gesture Recognition (FG 2020)(FG)}, pages 794--800.

\bibitem{kollias2019face}
Dimitrios Kollias, Viktoriia Sharmanska, and Stefanos Zafeiriou.
\newblock Face behavior a la carte: Expressions, affect and action units in a
  single network.
\newblock {\em arXiv preprint arXiv:1910.11111}, 2019.

\bibitem{kollias2021distribution}
Dimitrios Kollias, Viktoriia Sharmanska, and Stefanos Zafeiriou.
\newblock Distribution matching for heterogeneous multi-task learning: a
  large-scale face study.
\newblock {\em arXiv preprint arXiv:2105.03790}, 2021.

\bibitem{kollias2016line}
Dimitrios Kollias, Athanasios Tagaris, and Andreas Stafylopatis.
\newblock On line emotion detection using retrainable deep neural networks.
\newblock In {\em Computational Intelligence (SSCI), 2016 IEEE Symposium Series
  on}, pages 1--8. IEEE, 2016.

\bibitem{kollias2018deep1}
Dimitrios Kollias, Athanasios Tagaris, Andreas Stafylopatis, Stefanos Kollias,
  and Georgios Tagaris.
\newblock Deep neural architectures for prediction in healthcare.
\newblock {\em Complex \& Intelligent Systems}, 4(2):119--131, 2018.

\bibitem{kollias2019deep}
Dimitrios Kollias, Panagiotis Tzirakis, Mihalis~A Nicolaou, Athanasios
  Papaioannou, Guoying Zhao, Bj{\"o}rn Schuller, Irene Kotsia, and Stefanos
  Zafeiriou.
\newblock Deep affect prediction in-the-wild: Aff-wild database and challenge,
  deep architectures, and beyond.
\newblock {\em International Journal of Computer Vision}, pages 1--23, 2019.

\bibitem{kollias2020transparent}
Dimitris Kollias, Y Vlaxos, M Seferis, Ilianna Kollia, Levon Sukissian, James
  Wingate, and S Kollias.
\newblock Transparent adaptation in deep medical image diagnosis.
\newblock In {\em International Workshop on the Foundations of Trustworthy AI
  Integrating Learning, Optimization and Reasoning}, pages 251--267. Springer,
  2020.

\bibitem{kollias2017adaptation}
Dimitrios Kollias, Miao Yu, Athanasios Tagaris, Georgios Leontidis, Andreas
  Stafylopatis, and Stefanos Kollias.
\newblock Adaptation and contextualization of deep neural network models.
\newblock In {\em Computational Intelligence (SSCI), 2017 IEEE Symposium Series
  on}, pages 1--8. IEEE, 2017.

\bibitem{kollias2018aff}
Dimitrios Kollias and Stefanos Zafeiriou.
\newblock Aff-wild2: Extending the aff-wild database for affect recognition.
\newblock {\em arXiv preprint arXiv:1811.07770}, 2018.

\bibitem{kollias2018old}
Dimitrios Kollias and Stefanos Zafeiriou.
\newblock A multi-component cnn-rnn approach for dimensional emotion
  recognition in-the-wild.
\newblock {\em arXiv preprint arXiv:1805.01452}, 2018.

\bibitem{kollias2018multi}
Dimitrios Kollias and Stefanos Zafeiriou.
\newblock A multi-task learning \& generation framework: Valence-arousal,
  action units \& primary expressions.
\newblock {\em arXiv preprint arXiv:1811.07771}, 2018.

\bibitem{kollias2018training}
Dimitrios Kollias and Stefanos Zafeiriou.
\newblock Training deep neural networks with different datasets in-the-wild:
  The emotion recognition paradigm.
\newblock In {\em 2018 International Joint Conference on Neural Networks
  (IJCNN)}, pages 1--8. IEEE, 2018.

\bibitem{kollias2019expression}
Dimitrios Kollias and Stefanos Zafeiriou.
\newblock Expression, affect, action unit recognition: Aff-wild2, multi-task
  learning and arcface.
\newblock {\em arXiv preprint arXiv:1910.04855}, 2019.

\bibitem{kollias2020va}
Dimitrios Kollias and Stefanos Zafeiriou.
\newblock Va-stargan: Continuous affect generation.
\newblock In {\em International Conference on Advanced Concepts for Intelligent
  Vision Systems}, pages 227--238. Springer, 2020.

\bibitem{kollias2021affect}
Dimitrios Kollias and Stefanos Zafeiriou.
\newblock Affect analysis in-the-wild: Valence-arousal, expressions, action
  units and a unified framework.
\newblock {\em arXiv preprint arXiv:2103.15792}, 2021.

\bibitem{kollias2021analysing}
Dimitrios Kollias and Stefanos Zafeiriou.
\newblock Analysing affective behavior in the second abaw2 competition.
\newblock In {\em Proceedings of the IEEE/CVF International Conference on
  Computer Vision}, pages 3652--3660, 2021.

\bibitem{kollias2020exploiting}
Dimitrios Kollias and Stefanos~P Zafeiriou.
\newblock Exploiting multi-cnn features in cnn-rnn based dimensional emotion
  recognition on the omg in-the-wild dataset.
\newblock {\em IEEE Transactions on Affective Computing}, 2020.

\bibitem{latif2020augmenting}
Siddique Latif, Muhammad Asim, Rajib Rana, Sara Khalifa, Raja Jurdak, and
  Bj{\"o}rn~W Schuller.
\newblock Augmenting generative adversarial networks for speech emotion
  recognition.
\newblock {\em arXiv preprint arXiv:2005.08447}, 2020.

\bibitem{latif2019direct}
Siddique Latif, Rajib Rana, Sara Khalifa, Raja Jurdak, and Julien Epps.
\newblock Direct modelling of speech emotion from raw speech.
\newblock {\em arXiv preprint arXiv:1904.03833}, 2019.

\bibitem{latif2020deep}
Siddique Latif, Rajib Rana, Sara Khalifa, Raja Jurdak, and Bj{\"o}rn~W
  Schuller.
\newblock Deep architecture enhancing robustness to noise, adversarial attacks,
  and cross-corpus setting for speech emotion recognition.
\newblock {\em arXiv preprint arXiv:2005.08453}, 2020.

\bibitem{li2019reliable}
Shan Li and Weihong Deng.
\newblock Reliable crowdsourcing and deep locality-preserving learning for
  unconstrained facial expression recognition.
\newblock {\em IEEE Transactions on Image Processing}, 28(1):356--370, 2019.

\bibitem{li2017reliable}
Shan Li, Weihong Deng, and JunPing Du.
\newblock Reliable crowdsourcing and deep locality-preserving learning for
  expression recognition in the wild.
\newblock In {\em Proceedings of the IEEE conference on computer vision and
  pattern recognition}, pages 2852--2861, 2017.

\bibitem{mollahosseini2017affectnet}
Ali Mollahosseini, Behzad Hasani, and Mohammad~H Mahoor.
\newblock Affectnet: A database for facial expression, valence, and arousal
  computing in the wild.
\newblock {\em IEEE Transactions on Affective Computing}, 10(1):18--31, 2017.

\bibitem{pumarola2018ganimation}
Albert Pumarola, Antonio Agudo, Aleix~M Martinez, Alberto Sanfeliu, and
  Francesc Moreno-Noguer.
\newblock Ganimation: Anatomically-aware facial animation from a single image.
\newblock In {\em Proceedings of the European Conference on Computer Vision
  (ECCV)}, pages 818--833, 2018.

\bibitem{riaz2020exnet}
Muhammad~Naveed Riaz, Yao Shen, Muhammad Sohail, and Minyi Guo.
\newblock Exnet: An efficient approach for emotion recognition in the wild.
\newblock {\em Sensors}, 20(4):1087, 2020.

\bibitem{simonyan2014very}
Karen Simonyan and Andrew Zisserman.
\newblock Very deep convolutional networks for large-scale image recognition.
\newblock {\em arXiv preprint arXiv:1409.1556}, 2014.

\bibitem{srivastava2014dropout}
Nitish Srivastava, Geoffrey Hinton, Alex Krizhevsky, Ilya Sutskever, and Ruslan
  Salakhutdinov.
\newblock Dropout: a simple way to prevent neural networks from overfitting.
\newblock {\em The journal of machine learning research}, 15(1):1929--1958,
  2014.

\bibitem{sun2020mixup}
Lichao Sun, Congying Xia, Wenpeng Yin, Tingting Liang, Philip~S Yu, and Lifang
  He.
\newblock Mixup-transformer: dynamic data augmentation for nlp tasks.
\newblock {\em arXiv preprint arXiv:2010.02394}, 2020.

\bibitem{tagaris1}
Athanasios Tagaris, Dimitrios Kollias, and Andreas Stafylopatis.
\newblock Assessment of parkinson’s disease based on deep neural networks.
\newblock In {\em International Conference on Engineering Applications of
  Neural Networks}, pages 391--403. Springer, 2017.

\bibitem{tagaris2}
Athanasios Tagaris, Dimitrios Kollias, Andreas Stafylopatis, Georgios Tagaris,
  and Stefanos Kollias.
\newblock Machine learning for neurodegenerative disorder diagnosis—survey of
  practices and launch of benchmark dataset.
\newblock {\em International Journal on Artificial Intelligence Tools},
  27(03):1850011, 2018.

\bibitem{tan2019efficientnet}
Mingxing Tan and Quoc Le.
\newblock Efficientnet: Rethinking model scaling for convolutional neural
  networks.
\newblock In {\em International conference on machine learning}, pages
  6105--6114. PMLR, 2019.

\bibitem{thies2016face2face}
Justus Thies, Michael Zollhofer, Marc Stamminger, Christian Theobalt, and
  Matthias Nie{\ss}ner.
\newblock Face2face: Real-time face capture and reenactment of rgb videos.
\newblock In {\em Proceedings of the IEEE Conference on Computer Vision and
  Pattern Recognition}, pages 2387--2395, 2016.

\bibitem{wang2020region}
Kai Wang, Xiaojiang Peng, Jianfei Yang, Debin Meng, and Yu Qiao.
\newblock Region attention networks for pose and occlusion robust facial
  expression recognition.
\newblock {\em IEEE Transactions on Image Processing}, 29:4057--4069, 2020.

\bibitem{wei2018sample}
Shengyun Wei, Kele Xu, Dezhi Wang, Feifan Liao, Huaimin Wang, and Qiuqiang
  Kong.
\newblock Sample mixed-based data augmentation for domestic audio tagging.
\newblock {\em arXiv preprint arXiv:1808.03883}, 2018.

\bibitem{xu2018mixup}
Kele Xu, Dawei Feng, Haibo Mi, Boqing Zhu, Dezhi Wang, Lilun Zhang, Hengxing
  Cai, and Shuwen Liu.
\newblock Mixup-based acoustic scene classification using multi-channel
  convolutional neural network.
\newblock In {\em Pacific Rim conference on multimedia}, pages 14--23.
  Springer, 2018.

\bibitem{mdpi}
Miao Yu, Dimitrios Kollias, James Wingate, Niro Siriwardena, and Stefanos
  Kollias.
\newblock Machine learning for predictive modelling of ambulance calls.
\newblock {\em Electronics}, 10(4):482, 2021.

\bibitem{zafeiriou2017aff}
Stefanos Zafeiriou, Dimitrios Kollias, Mihalis~A Nicolaou, Athanasios
  Papaioannou, Guoying Zhao, and Irene Kotsia.
\newblock Aff-wild: Valence and arousal ‘in-the-wild’challenge.
\newblock In {\em Computer Vision and Pattern Recognition Workshops (CVPRW),
  2017 IEEE Conference on}, pages 1980--1987. IEEE, 2017.

\bibitem{zhang2017mixup}
Hongyi Zhang, Moustapha Cisse, Yann~N Dauphin, and David Lopez-Paz.
\newblock mixup: Beyond empirical risk minimization.
\newblock {\em arXiv preprint arXiv:1710.09412}, 2017.

\bibitem{zhou2017photorealistic}
Yuqian Zhou and Bertram~Emil Shi.
\newblock Photorealistic facial expression synthesis by the conditional
  difference adversarial autoencoder.
\newblock In {\em 2017 Seventh International Conference on Affective Computing
  and Intelligent Interaction (ACII)}, pages 370--376. IEEE, 2017.

\end{thebibliography}
}

\end{document}